\documentclass{ecai} 

\usepackage{latexsym}
\usepackage{amssymb}
\usepackage{amsmath}
\usepackage{amsthm}
\usepackage{booktabs}
\usepackage{enumitem}
\usepackage{graphicx}
\usepackage{color}
\usepackage{comment}
\usepackage{subfigure}
\usepackage{pgfplots}
\pgfplotsset{compat=1.18} 
\usepackage{multirow}
\usepackage{pifont}
\usepackage{subcaption}
\usepackage{hyperref}

\newcommand{\BibTeX}{B\kern-.05em{\sc i\kern-.025em b}\kern-.08em\TeX}
\newcommand{\link}[1]{{\color{blue}\ttfamily\href{#1}{#1}}}

\begin{document}


\begin{frontmatter}



\title{Context Matters: Leveraging Spatiotemporal Metadata for Semi-Supervised Learning on Remote Sensing Images}


\author[A]{\fnms{Maximilian}~\snm{Bernhard}\thanks{Corresponding Author. Email: bernhard@dbs.ifi.lmu.de}}
\author[A]{\fnms{Tanveer}~\snm{Hannan}}
\author[A]{\fnms{Niklas}~\snm{Strauß}}
\author[A]{\fnms{Matthias}~\snm{Schubert}} 

\address[A]{LMU Munich, MCML}

\begin{abstract}
Remote sensing projects typically generate large amounts of imagery that can be used to train powerful deep neural networks. However, the amount of labeled images is often small, as remote sensing applications generally require expert labelers. 
Thus, semi-supervised learning (SSL), i.e., learning with a small pool of labeled and a larger pool of unlabeled data, is particularly useful in this domain. 
Current SSL approaches generate pseudo-labels from model predictions for unlabeled samples. As the quality of these pseudo-labels is crucial for performance, utilizing additional information to improve pseudo-label quality yields a promising direction. 
For remote sensing images, geolocation and recording time are generally available and provide a valuable source of information as semantic concepts, such as land cover, are highly dependent on spatiotemporal context, e.g., due to seasonal effects and vegetation zones.
In this paper, we propose to exploit spatiotemporal metainformation in SSL to improve the quality of pseudo-labels and, therefore, the final model performance. We show that directly adding the available metadata to the input of the predictor at test time degenerates the prediction quality for metadata outside the spatiotemporal distribution of the training set. 
Thus, we propose a teacher-student SSL framework where only the teacher network uses metainformation to improve the quality of pseudo-labels on the training set.
Correspondingly, our student network benefits from the improved pseudo-labels but does not receive metadata as input, making it invariant to spatiotemporal shifts at test time.
Furthermore, we propose methods for encoding and injecting spatiotemporal information into the model and introduce a novel distillation mechanism to enhance the knowledge transfer between teacher and student.
Our framework dubbed \emph{Spatiotemporal SSL} can be easily combined with several state-of-the-art SSL methods, resulting in significant and consistent improvements on the BigEarthNet and EuroSAT benchmarks.
Code is available at \link{https://github.com/mxbh/spatiotemporal-ssl}.
\end{abstract}

\end{frontmatter}


\section{Introduction}
\begin{figure}[t]
    \centering
    \includegraphics[width=0.9\columnwidth]{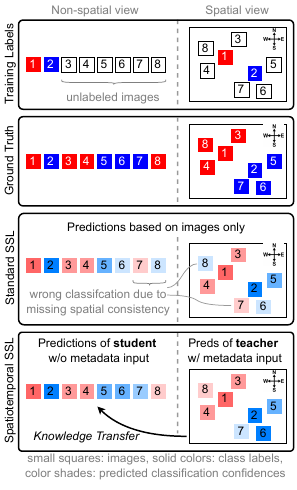}
    \caption{\textbf{Intuition for our proposed Spatiotemporal SSL framework.}
    Additional spatiotemporal input data facilitates learning for the teacher, leading to better predictions and pseudo-labels (7 and 8). The student model in Spatiotemporal SSL learns from these improved pseudo-labels without relying on the additional input, thereby also achieving a better performance.
    \emph{Temporal dimension omitted for simplicity. Best viewed digitally.}
    }
    \label{fig:teaser}
    \vspace{3mm}
\end{figure}
Applying deep learning models to analyze and interpret remote sensing imagery is a powerful tool that has been successfully applied in various use cases~\cite{han2023survey,yuan2020deep,cheng2020remote,osco2021review}.
However, deep neural networks are known to be data and label-hungry, i.e., they usually require large amounts of labeled samples to reach the desired performance~\cite{Goodfellow-et-al-2016}. This requirement poses a problem in many real-world applications as datasets with large quantities of high-quality labels matching the specific use case at hand are often unavailable, and labeling data is costly. In this regard, semi-supervised learning (SSL)~\cite{van2020survey} constitutes a promising remedy as it aims to narrow the performance gap to fully supervised learning when only a small part of the training dataset is labeled. 
While SSL is an active field of research with a multitude of methods and approaches~\cite{sohn2020fixmatch,lazarow2023unifying,chen2023softmatch,wang2023freematch,schmutz2023dont,xie2024class,lee2024cdmad}, its primary focus lies on optimizing the usage of unlabeled samples. At the same time, the potential of exploiting additional metadata is mostly neglected.

Remote sensing images are accompanied by metadata such as geolocation and acquisition time as they describe the Earth's surface at a certain place and time.
This spatiotemporal metadata can be used during training without additional labeling effort and improve the performance of models when used aptly~\cite{tang2015improving,minetto2019hydra,mac2019presence,chu2019geo,yang2022dynamic}. 
Location and recording time yield valuable information as many semantic concepts in remote sensing, such as land cover, are spatiotemporally coherent, and visual features are often highly dependent on the spatiotemporal context. For example, vegetation drastically varies with different climate zones, countries, and seasons.
Including this type of information is especially useful for SSL, where labeled data is scarce, as the additional features can alleviate the learning problem and, thus, ultimately reduce the label requirement~\cite{Goodfellow-et-al-2016}. However, current SSL methods do not include this valuable source of information.

In this paper, we propose an approach for leveraging this additional metainformation by learning the joint distribution of labels, visual features, and spatiotemporal context.
However, models using spatiotemporal metadata as an input are prone to overfitting to the metadata of the training set due to the small number of labeled samples in SSL and spatiotemporal sampling biases introduced during dataset creation. 
As a result, models relying on spatiotemporal metadata as an input poorly generalize at test time for out-of-distribution metadata (see experiments in Section~\ref{sec:ood}). 
Especially in remote sensing, datasets often cover only a limited area and a few points in time. Therefore, applying a model in spatiotemporal contexts beyond the training data, e.g., a different geographical region or season, is highly desirable in practice.

To leverage spatiotemporal metainformation in SSL without compromising generalization at test time, we propose a novel student-teacher framework, which we call \emph{Spatiotemporal SSL} (see Figure~\ref{fig:teaser}). 
The teacher model receives the metadata as additional input and is trained in a semi-supervised way to generate high-quality pseudo-labels. These pseudo-labels are then used to train the student model, which does not rely on the metadata input but solely operates on images. 
This design has the following advantages: 
First, the teacher model benefiting from the spatiotemporal information is only employed on the training set, avoiding the generalization issues at inference mentioned above. 
Second, the student model indirectly benefits from the spatiotemporal information as it receives strong pseudo-labels generated with the help of the additional input. 
Third, since the student model does not receive the metadata input, it is invariant to shifts in the spatiotemporal distribution of test samples, allowing it to generalize to unseen spatiotemporal contexts (see experiments in Section~\ref{sec:ood}).

To jointly model visual features and the spatiotemporal distribution in the teacher, we modify the vision transformer (ViT)~\cite{dosovitskiy2021an} by supplementing the visual patch tokens with a specialized metatoken encoding the spatiotemporal information. Furthermore, we optimize the knowledge transfer between teacher and student beyond passing on pseudo-labels by integrating a novel distillation mechanism. Here, a dedicated distillation token in the student model is supervised to align with the metatoken embedding of the teacher model, allowing the student model to access the spatiotemporal reasoning of the teacher without actually receiving spatiotemporal inputs.

Notably, our Spatiotemporal SSL framework does not rely on restrictive assumptions about the underlying SSL algorithm, making it versatile and compatible with recent developments in SSL. In our experiments on the popular BigEarthNet and EuroSAT benchmarks, we demonstrate that combining Spatiotemporal SSL with several state-of-the-art SSL methods such as FixMatch~\cite{sohn2020fixmatch} and DeFixMatch~\cite{schmutz2023dont} leads to substantial and consistent improvements. We also perform detailed experiments and ablation studies to identify and analyze relevant factors in our approach.

\section{Related Work}
\paragraph{Semi-Supervised Learning}
Semi-supervised learning (SSL) is an active field of research with a large variety of methods~\cite{van2020survey,yang2022survey,ouali2020overview}. A major cornerstone in this field is FixMatch~\cite{sohn2020fixmatch}. FixMatch represents a framework where hard pseudo-labels are generated from confident predictions on weakly augmented, unlabeled samples in order to supervise the predictions for strongly augmented versions of these samples. 
Many works extend FixMatch, e.g., by applying specialized consistency losses~\cite{fan2023revisiting,li2021comatch,zheng2022simmatch,zheng2023simmatchv2,schmutz2023dont} or replacing the hard, threshold-based pseudo-labeling with more sophisticated techniques, oftentimes aiming to align the distribution of pseudo-labels with the observed distribution of labels~\cite{zhang2021flexmatch,xu2021dash,berthelot2021adamatch,wang2023freematch,chen2023softmatch}.
More specifically, FreeMatch~\cite{wang2023freematch} adapts FixMatch's fixed confidence threshold for pseudo-labeling for each class separately and combats pseudo-label class imbalance with an additional loss.
SoftMatch~\cite{chen2023softmatch} applies a soft pseudo-label weighting and, similarly to FreeMatch, aligns the pseudo-labels with a uniform distribution. DeFixMatch~\cite{schmutz2023dont} follows a rather different approach, which is debiasing the learner by applying the negative unsupervised loss on the labeled data. 

Many works in SSL focus on problems that are often met when working with remote sensing data.
For example,~\cite{wei2021crest,peng2023dynamic,lazarow2023unifying} consider SSL with imbalanced and long-tailed data. In particular, UDAL~\cite{lazarow2023unifying} presents a way to combat label imbalance by integrating the distribution alignment into the cross-entropy computation via modulating predicted scores. CAP~\cite{xie2024class} addresses the problem of semi-supervised multi-label classification (instead of multi-class, single-label classification as most SSL benchmarks). Their approach is centered around finding suitable pseudo-labeling thresholds for the individual classes.

Our contribution can be considered orthogonal to previous methods since our framework can be combined with several state-of-the-art SSL algorithms as shown in our experiments. Hence, even future SSL techniques might benefit from our approach.

\paragraph{Image Classification with Additional Metadata}
Several works explored feeding image metadata as additional input to the model in fully supervised settings.
In~\cite{tang2015improving}, it has been shown that geolocation as additional network input can improve supervised image classification on YFCC100M~\cite{thomee2016yfcc100m}. 
Similarly, \cite{minetto2019hydra} uses metadata, including geolocation, as additional input in supervised classification on FMOW~\cite{christie2018functional}. However, their main focus lies on network ensembling technique instead of the inclusion of additional input data.
In~\cite{salem2020learning}, metadata are used via context networks to learn a dynamic map of visual appearance, which has been shown to be beneficial for image localization, image retrieval, and metadata verification.
\cite{mac2019presence} exploits metadata to learn a spatiotemporal prior in order to refine predictions in supervised fine-grained classification on YFCC100M~\cite{thomee2016yfcc100m}, BirdSnap~\cite{berg2014birdsnap}, and iNaturalist~\cite{van2018inaturalist}. Especially for fine-grained image classification, the usage of spatiotemporal metadata has proven useful as the subsequent works of~\cite{chu2019geo} and~\cite{yang2022dynamic} demonstrate. Specifically,~\cite{chu2019geo} investigates different ways to include metadata, i.e., via geographical priors, prediction postprocessing, or feature modulation, whereas~\cite{yang2022dynamic} proposes DynamicMLP, a novel building block for fusing multimodal features effectively.

Though existing literature shows the value of spatiotemporal metadata for image classification, this line of research only examines fully supervised settings. 
Furthermore, these works do not address spatiotemporal generalization at test time. 
In contrast, we examine spatiotemporal metadata in SSL where directly using the metadata as input might lead to generalization issues.

\section{Spatiotemporal SSL}
\begin{figure*}[t]
    \centering
    \includegraphics[width=0.99\textwidth]{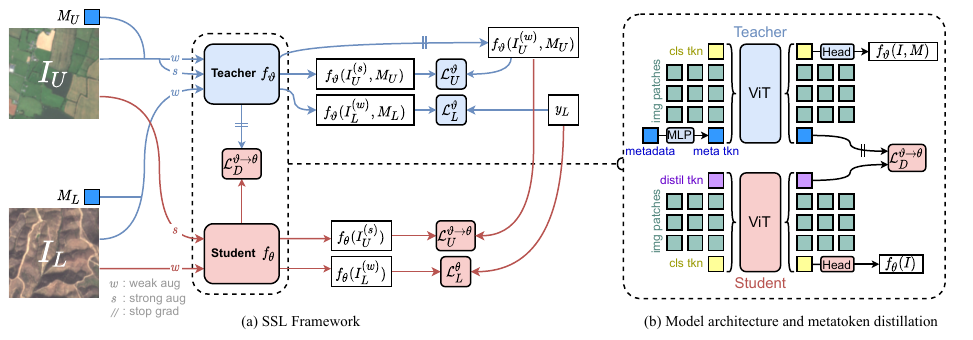}
    \caption{\textbf{Overview of the proposed Spatiotemporal Semi-supervised Learning.}
        The spatiotemporal teacher $f_\vartheta$ utilizes both images and metadata and generates strong pseudo-labels that are used as supervision for the student $f_\theta$, which does not access the metadata. 
    }
    \label{fig:overall}
    \vspace{3mm}
\end{figure*}
In the following, we introduce the necessary notation before we describe the training and architecture of the spatiotemporal teacher model, the student's training, and the novel distillation scheme. Figure~\ref{fig:overall} provides an overview of our approach.

\subsection{Notation}
Let $I$ and $y$ denote images and their ground-truth labels. In SSL, we further distinguish between labeled and unlabeled images $I_L$ and $I_U$. We denote the number of labeled images and unlabeled images in a batch as $n_L$ and $n_U$, respectively.
During training, we only have access to the ground truth of the labeled images $y_L$.
We consider the geolocation $G$ and time $T$ of image acquisition as additional metadata, consolidated to $M=(G,T)$ ($M_L$ and $M_U$ for labeled and unlabeled samples, respectively). 
Our ultimate goal is to estimate the label $y$ for an image $I$ with a neural network $f_\theta$ (student) as $f_\theta(I)$. The metadata $M$ will be consumed by another neural network $f_\vartheta$ (teacher), providing the pseudo-label $f_\vartheta(I,M)$. 
We assume $f_\theta$ and $f_\vartheta$ to be two separate neural networks as their input differs.

\subsection{Training the Spatiotemporal Teacher Model}
To train the spatiotemporal teacher model, we assume a generic SSL framework. In the following, we adopt the basic structure of FixMatch~\cite{sohn2020fixmatch}, which is the basis for most recent SSL methods.
Here, the teacher model $f_\vartheta$ is trained with separate losses for labeled and unlabeled data. 
The loss for the labeled data $\mathcal{L}_L$ can be any supervised loss function, i.e., we write
\begin{align}
    \mathcal{L}_L^{\vartheta} = \frac{1}{n_L} \sum_{i=0}^{n_L} \ell_L (y_i, f_\vartheta(I_i,M_i))
    \label{eq:ll}
\end{align}
where $\ell_L$ denotes the loss function for a single sample and is chosen according to the underlying SSL algorithm (e.g., cross-entropy in FixMatch).
On the other hand, the unsupervised loss $\mathcal{L}_U^{\vartheta}$ is applied to ensure consistency on unlabeled data. This loss employs pseudo-labels generated from weakly augmented image versions $I_U^{(w)}$ to supervise the predictions for strongly augmented image versions $I_U^{(s)}$. To mitigate the effect of noisy and unreliable pseudo-labels, a weighting function $\alpha$ determines the loss contribution of each individual sample (e.g., confidence-based hard thresholding as in FixMatch~\cite{sohn2020fixmatch}). Formally, the unlabeled loss $\mathcal{L}_U$ is defined as
\begin{align}
    \mathcal{L}_U^{\vartheta} = \frac{1}{n_U}\!\sum_{i=1}^{n_U}\!\alpha\!\left(f_\vartheta(I_i^{(w)}\!, M_i)\right) \cdot \ell_U\! \left(f_\vartheta(I_i^{(w)}\!,M_i), f_\vartheta(I_i^{(s)}\!,M_i) \right)\!.
    \label{eq:lu}
\end{align}
For example, we can use cross-entropy as criterion $\ell_U$ after generating hard pseudo-labels from $f_\vartheta(I_i^{(w)}, M_i)$. Let us note that more sophisticated techniques, such as modulating prediction scores with distribution alignment \cite{wang2023freematch,chen2023softmatch}, can also be used.
The overall loss for the teacher model $f_\vartheta$ is
\begin{align}
    \mathcal{L}^{\vartheta} = \mathcal{L}_L^{\vartheta} + \lambda_U \cdot \mathcal{L}_U^{\vartheta},
    \label{eq:lt}
\end{align}
where $\lambda_U$ is a parameter for balancing the two described loss terms.

\subsection{Spatiotemporal Teacher Architecture}
A straightforward way to exploit spatiotemporal metadata would be to estimate class priors from the spatiotemporal occurrences and use these to post-process the predicted classification scores, e.g., with a Bayesian approach~\cite{chu2019geo,mac2019presence}. 
This approach assumes that the visual appearance of a class is generally the same for different geolocations and times. More formally, the assumption is that the visual appearance $I|y$ given the class is conditionally independent of the spatiotemporal context $M$~\cite{mac2019presence}.
However, in our setting, we argue that the spatiotemporal context has a considerable effect on the visual appearance as, for instance, the same type of land cover may look very different across varying countries, climate zones, and seasons. 

Therefore, it is sensible to process and modulate image features conditional to the spatiotemporal information as it allows the model to capture such variation in the visual appearance. To put this into practice, we opt for an early-fusion architecture where spatiotemporal and image information are modeled jointly.
In other words, with an early interaction of visual and spatiotemporal information, the model can not only learn which classes are likely for a certain location and time but also how they visually depend on the location and time.

On a technical level, we concatenate latitude, longitude, and time, represented as the relative day of the year, and feed the resulting vector into a two-layer MLP to generate a single \emph{metatoken}. 
Afterward, we pass the metatoken along with the visual tokens encoding image patches on to the Vision Transformer (ViT)~\cite{dosovitskiy2021an} architecture (see the teacher in Figure~\ref{fig:overall}~(b)). This design allows maximal interaction between the visual and spatiotemporal information while introducing minimal methodological overhead. In addition, the metatoken can be easily injected into other transformer-based vision backbones such as~\cite{fan2021multiscale,cong2022satmae,prithvi-100M-preprint,reed2023scale}. An empirical analysis of the proposed early-fusion approach can be found in Section~\ref{sec:ablations}.

\subsection{Training the Student Model}
Building upon the training of the teacher model $f_\vartheta$, we now describe the training of the student $f_\theta$. Our goal is to transfer the teacher's learned knowledge to the student not receiving metadata as input. To this end, we train $f_\theta$ with its own loss for labeled data $\mathcal{L}_L^{\theta}$, which can be obtained from  Equation~\ref{eq:ll} by substituting $f_\theta(I_i)$ for $f_\vartheta(I_i,M_i)$. 
However, we want the student to benefit from the spatiotemporal teacher by providing it with the teacher's pseudo-labels for unlabeled data. More precisely, we modify $\mathcal{L}_U$ from Equation~\ref{eq:lu} as follows:
\begin{align}
    \mathcal{L}_U^{\vartheta\rightarrow\theta} = \frac{1}{n_U}\!\sum_{i=1}^{n_U}\!\alpha\!\left(f_\vartheta(I_i^{(w)}\!,M_i)\right) \cdot \ell_U\!\left(f_\vartheta(I_i^{(w)}\!, M_i), f_\theta(I_i^{(s)}) \right)\!.
    \label{eq:lus}
\end{align}
That is, instead of bootstrapping the pseudo-labels for $f_\theta$ from $f_\theta$'s predictions itself, we employ the pseudo-labels generated from the teacher $f_\vartheta$. As $f_\vartheta$ has access to the additional input, these pseudo-labels are of higher quality and, therefore, improve the student's training.
Putting everything together, we obtain the training objective for $f_\theta$ as 
\begin{align}
    \mathcal{L}^{\theta} = \mathcal{L}_L^{\theta} + \lambda_U \cdot \mathcal{L}_U^{\vartheta\rightarrow\theta}.
    \label{eq:ls}
\end{align}
The teacher and the student are trained simultaneously on the same images, resulting in convenient single-stage training.

Let us note that Spatiotemporal SSL does not make restrictive assumptions about the underlying SSL algorithm, i.e., $\alpha$, $\ell_L$, $\ell_U$ as well as additional loss functions such as the debiasing loss of \cite{schmutz2023dont} can be freely chosen. 
Hence, our approach is versatile and orthogonal to recent developments in SSL, allowing to combine it with several state-of-the-art SSL algorithms (see Section~\ref{sec:sota}).

\subsection{Metatoken Distillation}
To enhance the knowledge transfer from the teacher to the student beyond the exchange of pseudo-labels, we integrate a specialized knowledge distillation mechanism into our framework (see Figure~\ref{fig:overall}~(b)). 
Inspired by~\cite{touvron2021training}, we introduce a dedicated distillation token to the student network, which interacts with all the other tokens through self-attention. 
In contrast to~\cite{touvron2021training}, we do not use this token for distilling knowledge from another model with a fundamentally different architecture but from a model with different inputs.
That is, the output embedding of the distillation token is supervised to be similar to the output embedding of the teacher's spatiotemporally informed metatoken. 
Our proposed metatoken distillation loss $\mathcal{L}^{\vartheta\rightarrow\theta}_D$ is defined as the mean squared error of the teacher's metatoken embeddings and the student's distillation token embeddings over all samples (labeled and unlabeled). 
We use this loss to update only the student model $f_\theta$ but not the teacher $f_\vartheta$ to prevent the teacher from adapting to the student. Since the student does not access the additional metainformation, it cannot provide meaningful guidance for the teacher (see Section~\ref{sec:ablations} for an empirical justification). 
Hence, the overall loss for the student $f_\theta$ is defined as 
\begin{align}
    \mathcal{L}^{\theta} = \mathcal{L}_L^{\theta} + \lambda_U \cdot \mathcal{L}_U^{\vartheta\rightarrow\theta} + \lambda_D \cdot \mathcal{L}_D^{\vartheta\rightarrow\theta},
\end{align}
where $\lambda_D$ is a weighting hyperparameter for the distillation loss.
A detailed analysis of design choices and ablation studies for this additional loss can be found in Section~\ref{sec:ablations}.

\section{Experiments}
\subsection{Datasets}
\paragraph{BigEarthNet}
\cite{sumbul2019bigearthnet} is a large-scale land cover dataset. It contains 590,326 Sentinel-2 images of 120$\times$120 pixels and 43 classes taken from the CORINE Land Cover database (CLC18). 
BigEarthNet is an imbalanced multi-label classification dataset, where the relative frequencies of labels ranges from 37\% to 0.05\%.
We adopt the train-val-test split of~\cite{manas2021seasonal} and further split the training set into labeled and unlabeled samples in a stratified way. Apart from the RGB-bands of the images, we consider the geolocation and the image acquisition time, represented as the relative day of the year, as the input to the network. The BigEarthnet images were acquired between June 2017 and May 2018 in ten European countries (Austria, Belgium, Finland, Ireland, Kosovo, Lithuania, Luxembourg, Portugal, Serbia, Switzerland).

\paragraph{EuroSAT}
\cite{helber2019eurosat} is another, highly popular land use and land cover dataset containing 27,000 Sentinel-2 patches of 64$\times$64 pixels. Each image patch belongs to one out of 10 classes. We adopt the dataset split provided by the Unified Semi-supervised Learning Benchmark (USB)~\cite{usb2022}, i.e., we use a fixed train-test split and the training images are further divided into labeled and unlabeled images such that exactly the same number of labels are available for every class. Similar to BigEarthNet, we use the RGB-bands of the images and the geolocation as metadata but omit the image acquisition time, which is not available for EuroSAT. The image locations of EuroSAT are scattered over 34 European countries (see supplementary material\footnote{supplementary material can be found in our code repository}).

\subsection{Implementation Details}
We implement our method within the USB codebase~\cite{usb2022}. We use ViT-S~\cite{dosovitskiy2021an} as our base architecture. On BigEarthNet, we resize images to 128$\times$128, use a ViT patch size of 16 pixels, and train with a cosine learning rate schedule for 64k steps with an initial learning rate of 1e-4 and a batch size of 512 (64 labeled, 448 unlabeled). 
On EuroSAT, we adopt the configuration of USB~\cite{usb2022}, i.e., we resize images to 32$\times$32 pixels, use a ViT patch size of 2 pixels, and train with a cosine learning rate schedule for 204,800 steps with an initial learning rate of 5e-5 and a batch size of 16 (8 labeled, 8 unlabeled). 
As BigEarthNet is a multi-label classification dataset, we extend SSL methods that were originally proposed for single-label problems (FixMatch~\cite{sohn2020fixmatch}, SoftMatch~\cite{chen2023softmatch}, FreeMatch~\cite{wang2023freematch}, DeFixMatch~\cite{schmutz2023dont}, UDAL~\cite{lazarow2023unifying}) by applying the pseudo-labeling scheme to every class separately.
For the weight parameter $\lambda_U$, we adopt the proposed values of the respective underlying SSL algorithm. For $\lambda_D$, we choose 1.0 on BigEarthNet and 0.01 on EuroSAT (see Section~\ref{sec:ablations}).
Following the common practice~\cite{sohn2020fixmatch}, we use the exponential moving average of models for evaluation. 
For more details, see supplementary material.

\subsection{Combination and Comparison with State-of-the-Art SSL Methods on BigEarthNet}
\label{sec:sota}
\begin{table}[t]
    \centering
    \caption{\textbf{Results on BigEarthNet with 1\% training labels} Our ST-SSL leads to an improvement on every standard SSL method, even surpassing the best standard algorithm (DeFixMatch) with the worst ST-SSL combination (CDMAD + ST-SSL).}
    \begin{tabular}{rlcc}
    \toprule
    \textbf{Method} & \textit{(Venue)} & \textbf{mAP} & \textit{(}$\pm\Delta$\textit{)}\\ \midrule
    \multicolumn{2}{c}{Supervised only baseline} & 40.47 & -- \\ \midrule
    FixMatch~\cite{sohn2020fixmatch} &(\textit{NeurIPS'20}) & 42.56 & --\\
    UDAL~\cite{lazarow2023unifying}  &(\textit{WACV'23})    & 42.32 & --\\
    SoftMatch~\cite{chen2023softmatch} &(\textit{ICLR'23})  & 41.95 & --\\
    FreeMatch~\cite{wang2023freematch} &(\textit{ICLR'23})  & 42.46 & --\\
    DeFixMatch~\cite{schmutz2023dont} &(\textit{ICLR'23})   & \underline{43.09} & --\\
    CAP~\cite{xie2024class} &(\textit{NeurIPS'23})          & 41.81 & --\\ 
    CDMAD~\cite{lee2024cdmad}&(\textit{CVPR'2024})          & 41.48 & --\\
    \midrule
    FixMatch\quad +& ST-SSL   & 46.12 & \textit{(+3.56)} \\
    UDAL\quad +& ST-SSL       & 45.84 & \textit{(+3.52)} \\
    Softmatch\quad +& ST-SSL  & 45.34 & \textit{(+3.39)} \\
    FreeMatch\quad +& ST-SSL  & 45.20 & \textit{(+2.74)} \\
    DeFixMatch\quad +& ST-SSL & \textbf{46.65} & \textit{(+3.56)} \\
    CAP\quad +& ST-SSL        & 45.53 & \textit{(+3.72)} \\
    CDMAD\quad+& ST-SSL       & 43.74 & \textit{(+2.26)} \\
    \bottomrule
    \end{tabular}
    \label{tab:sota}
\end{table}
\begin{figure}[t]
    \centering
    \begin{tikzpicture}
    \begin{axis}[
        width=\columnwidth,
        height=49mm,
        xlabel={Percentage of training labels},
        ylabel={mAP},
        xmin=0, xmax=8.5,
        ymin=34, ymax=61,
        xtick={0.5, 1, 2, 4, 8},
        ytick={35,40,45,50,55,60,65},
        legend pos=south east,
        ymajorgrids=true,
        xmajorgrids=true,
        grid style=dashed,
    ]
    \addplot[
        color=black!50!green,
        mark=*,
        mark options={scale=0.8}
        ]
        coordinates {
        (0.5,40.65)(1,46.12)(2,51.65)(4,54.80)(8,59.90)
        };
        \addlegendentry{FixMatch + ST-SSL}
    
    \addplot[
        color=red,
        mark=triangle*,
        mark options={scale=0.8}
        ]
        coordinates {
        (0.5,39.27)(1,42.56)(2,48.49)(4,52.99)(8,57.48)
        };
        \addlegendentry{FixMatch}
        
    \addplot[
        color=blue,
        mark=square*,
        mark options={scale=0.8}
        ]
        coordinates {
        (0.5,36.23)(1,40.47)(2,47.51)(4,51.43)(8,55.78)
        };
        \addlegendentry{supervised only}
        
    \end{axis}
    \end{tikzpicture}
    \caption{\textbf{Results for different numbers of labeled training samples on BigEarthNet.} Our method consistently outperforms FixMatch and the supervised baseline.}
    \label{fig:curves}
    \vspace{5mm}
\end{figure}
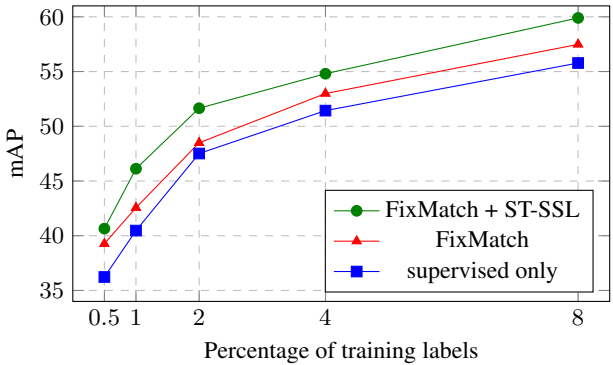

In Table~\ref{tab:sota}, we provide results for several SSL methods on BigEarthNet when 1\% of the training labels are available. The existing SSL methods reach mAP scores in the range of 41.48\% (CDMAD) to 43.09\% (DeFixMatch), thereby surpassing the "supervised only baseline" (40.47\%), which is trained in a fully supervised way on only the labeled samples. When combining these SSL methods with the Spatiotemporal SSL framework, we can observe consistent performance improvements leading to mAP scores in the range of 43.74\% (CDMAD + ST-SSL) and 46.65\% (DeFiXMatch + ST-SSL). 
That is, even the least performing ST-SSL variant outperforms the best SSL method without ST-SSL.
At the same time, the top-performing conventional SSL method DeFixMatch improves by 3.56\% mAP with ST-SSL. This observation is even more remarkable since we did not tune hyperparameters for the single ST-SSL combinations, but took the conventional SSL configuration and transferred it to ST-SSL without modification.

Furthermore, we investigate the generalization of our approach to other settings with different proportions of labeled data by comparing FixMatch and FixMatch + ST-SSL in Figure~\ref{fig:curves}. Once again, ST-SSL consistently improves the baseline across all settings.

\subsection{Ablations}
\label{sec:ablations}
\begin{table}[t]
    \centering
    \caption{\textbf{Ablations and detailed experiments} on BigEarthNet with 1\% training labels.}
    \begin{tabular}{clcc}
    \toprule
     & \textbf{Method}  & \textbf{mAP} & \textit{(}$\pm\Delta$\textit{)} \\ \midrule   
    (a) & \begin{tabular}{@{}l} FixMatch + ST-SSL \\ \hspace{5mm}(MSE distil, $\lambda_D=1$)\end{tabular}   & 46.12 & -- \\ \midrule
    (b) & without acquisition time $T$   & 45.59 & \textit{(-0.53)} \\ 
    (c) & without geolocation $G$       & 44.96 & \textit{(-1.16)} \\ \midrule
    (d) & late fusion                & 43.59 & \textit{(-2.53)}\\
    (e) & single model               & 44.61 & \textit{(-1.51)} \\ \midrule
    (f) & without distillation       & 44.92 & \textit{(-1.20)} \\ 
    (g) & MAE distillation           & 45.86 & \textit{(-0.36)} \\ 
    (h) & cosine sim.~distillation   & 45.17 & \textit{(-0.95)} \\ 
    (i) & classification token distillation   & 42.89 & \textit{(-3.23)} \\
    (j) & without stopping gradients & 45.86 & \textit{(-0.26)} \\ \midrule
    (k) & $\lambda_D = 0.1$          & 45.27 & \textit{(-0.85)} \\
    (l) & $\lambda_D = 0.5$          & 45.76 & \textit{(-0.36)} \\
    (m) & $\lambda_D = 2.0$          & 45.78 & \textit{(-0.34)} \\
    \bottomrule
    \end{tabular}
    \label{tab:ablation}
\end{table}
To identify contributing factors within our method, we perform a variety of ablation studies presented in Table~\ref{tab:ablation}. We base the ablations on FixMatch~\cite{sohn2020fixmatch} as it is a strong and highly popular SSL method that is conceptually simple at the same time. 

Table~\ref{tab:ablation} is divided into five sections: The first section (a) represents our full method as already seen in Table~\ref{tab:sota}. 
Next, experiments (b,c), analyze the effect of geolocation $G$ and acquisition time $T$. If we omit the acquisition time $T$ and only consider the geolocation $G$ as additional input (b), we observe a drop in mAP of about 1\%. Conversely, if we only use the image acquisition time and omit the geolocation (c), we observe a performance drop of about 2\%. That is, both the geolocation and acquisition time provide useful information for the model. However, the geolocation seems to have a stronger effect which is in line with the reasoning that land cover is geographically coherent. On the other hand, the time information is helpful as it allows the network to model differences in visual appearance of land cover throughout seasons and potentially also because of sampling bias in the training data. Note that exploiting sampling bias in the training data is not problematic in our framework as the final model has no direct access to the metadata, preventing it from transferring the bias to test time.

The third group (d,e) in Table~\ref{tab:ablation} is concerned with the modeling of spatiotemporal information within the teacher network. First, we compare our early fusion via the metatoken with a late-fusion approach (d), where we add the encoded metadata to the encoded image representation before passing it on to the classification head. 
The resulting performance degradation of about 2.5\% mAP indicates that the early interaction of image features and metadata information is, in fact, beneficial as it allows the modulation of visual features based on the spatiotemporal context. 
In the late-fusion architecture, this kind of interaction is not possible, restricting the model to primarily learn a spatiotemporal prior for the occurrence of classes. Furthermore, we investigate the case where the teacher and the student are represented by the same network (e). Here, we observe a decrease in performance of 1.51\%, indicating that specialization with separate models for spatiotemporal and purely visual reasoning is beneficial.

In the next section (f-j), we investigate several design choices regarding our metatoken distillation. First of all, not using the distillation mechanism at all (f), leads to an inferior performance of 44.92\% mAP (vs.~46.12\%). Replacing the mean squared error (MSE) in the distillation loss $\mathcal{L}_D^{\vartheta \rightarrow \theta}$ with the mean absolute error (MAE) (g) and the cosine similarity (h) results in mAP drops of 0.36\% and 0.95\%, respectively. Furthermore, using the teacher's classification token instead of the metatoken for distillation (i) leads to a drop of 3.23\% mAP. Moreover, omitting the gradient stopping and letting the gradients of the distillation loss flow into the teacher network (j) decreases the performance by a slight margin of 0.26\%. 

Finally, we study the influence of the weight of the distillation loss $\lambda_D$ in the last section of the table (k-m). 
The overall performance is relatively robust to the choice of $\lambda_D$ as setting $\lambda_D$ suboptimally still gives us better mAP scores than not using the metatoken distillation at all (f). For an analogous experiment on EuroSAT, please refer to the supplementary material.

\begin{figure*}[t]
    \centering
    \includegraphics[width=\textwidth]{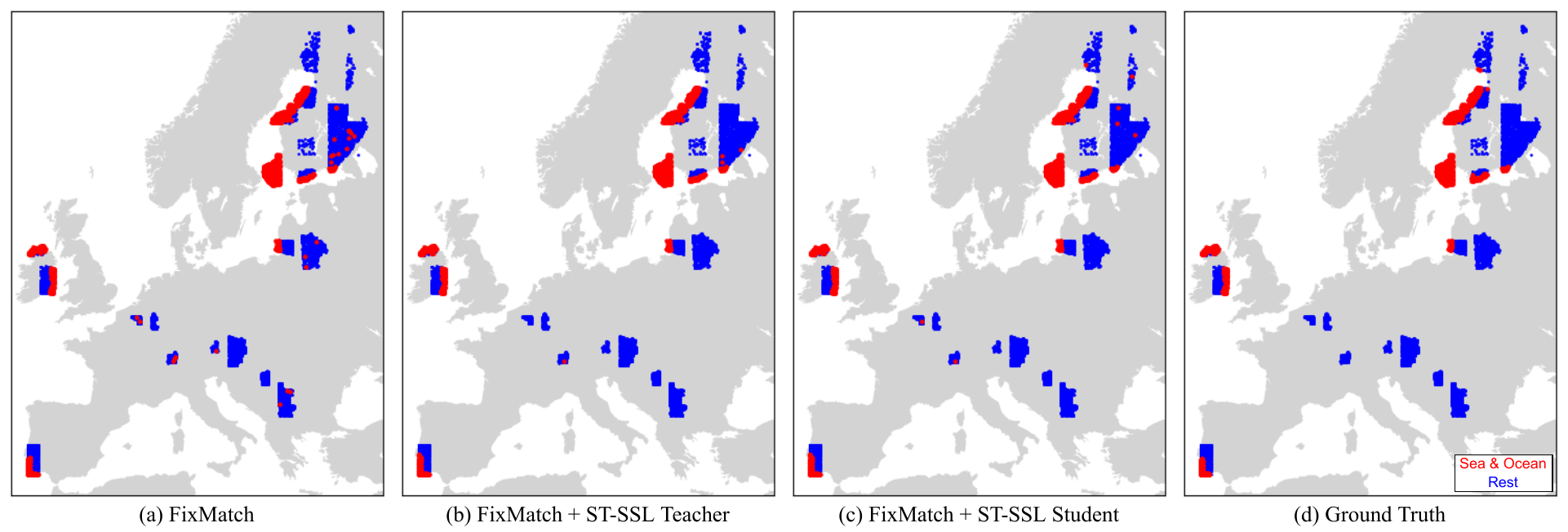}
    \caption{
        \textbf{Qualitative comparison of FixMatch and our method} on BigEarthNet. 
        As FixMatch (a) does not exploit the geospatial context of images, it produces numerous false positive predictions for the class "sea \& ocean" in mainland regions (e.g., Eastern Finland). In contrast, our spatiotemporal teacher model (b) is able to largely mitigate this type of error by considering the image geolocation, leading to more accurate pseudo-labels, from which the student model (c) benefits as well. \emph{Best viewed digitally.}
    }
    \label{fig:qual}
\end{figure*}
\begin{figure}[t]
    \centering
    \includegraphics[width=1.0\columnwidth]{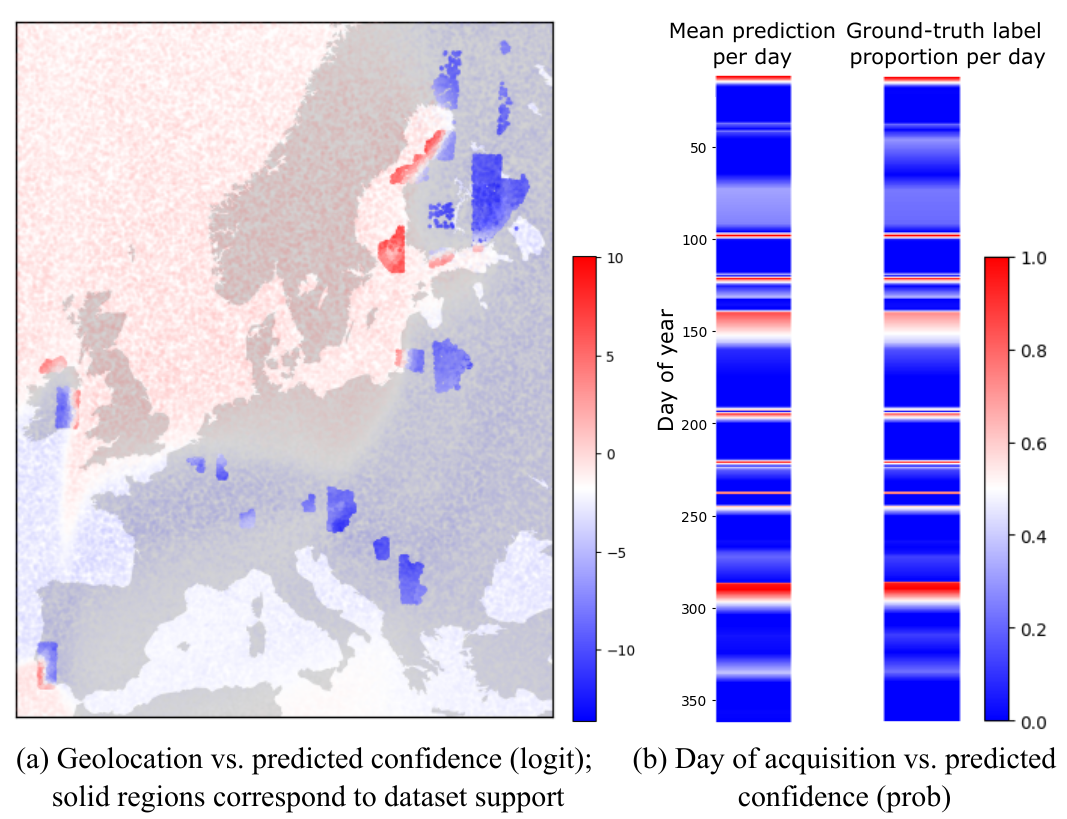}
    \caption{
        \textbf{Visualizing the spatiotemporal prior learned by the teacher model}.
        We feed a monochrome, gray image together with all metadata instances of BigEarthNet into the model and visualize the confidence for the class "sea \& ocean" depending on geolocation (a) and time of image acquisition (b). We can see that the model has learned the dataset's spatiotemporal distribution well (see Figure~\ref{fig:qual}~(d) for geolocation vs.~ground-truth label) but does not generalize to OOD locations.
        \emph{Best viewed digitally.}
    }
    \label{fig:all_meta}
    \vspace{5mm}
\end{figure}


\subsection{Generalization to Out-of-Distribution Metadata}
\label{sec:ood}
\begin{table}[t]
    \centering
    \caption{\textbf{Generalization to out-of-distribution metadata} on BigEarthNet with 1\% training labels. $\dagger$: Different dataset split based on geography, i.e., (e,f) are \emph{not} directly comparable to (a-d).}
    \begin{tabular}{c@{\hspace{1mm}}c@{\hspace{1mm}}c@{\hspace{0mm}}c@{\hspace{0mm}}c}
        \toprule
        & \textbf{Model} & \textbf{Metadata} &\textbf{OOD component} & \textbf{mAP} \\ \midrule
        (a) & Student & \ding{55} & -- & 46.12     \\
        (b) & Teacher & \ding{51} & -- & 44.93     \\
        \midrule
        (c) & Teacher & \ding{51} & $p_{train}(G) \neq p_{test}(G)$ & 30.15 $\pm$ 1.38 \\
        (d) & Teacher & \ding{51} & $p_{train}(T) \neq p_{test}(T)$ & 42.75 $\pm$ 1.30 \\ \midrule
        (e) & Student$^\dagger$ & \ding{55} & $p_{train}(y,I) \neq p_{test}(y,I)$ & 16.65 \\
        (f) & Teacher$^\dagger$ & \ding{51} & $p_{train}(y,I,G) \neq p_{test}(y,I,G)$ & 15.43 \\
        \bottomrule
        \
    \end{tabular}
    \label{tab:ood}
    \vspace{0mm}
\end{table}
\begin{figure}[t]
    \centering
    \includegraphics[width=1.0\columnwidth]{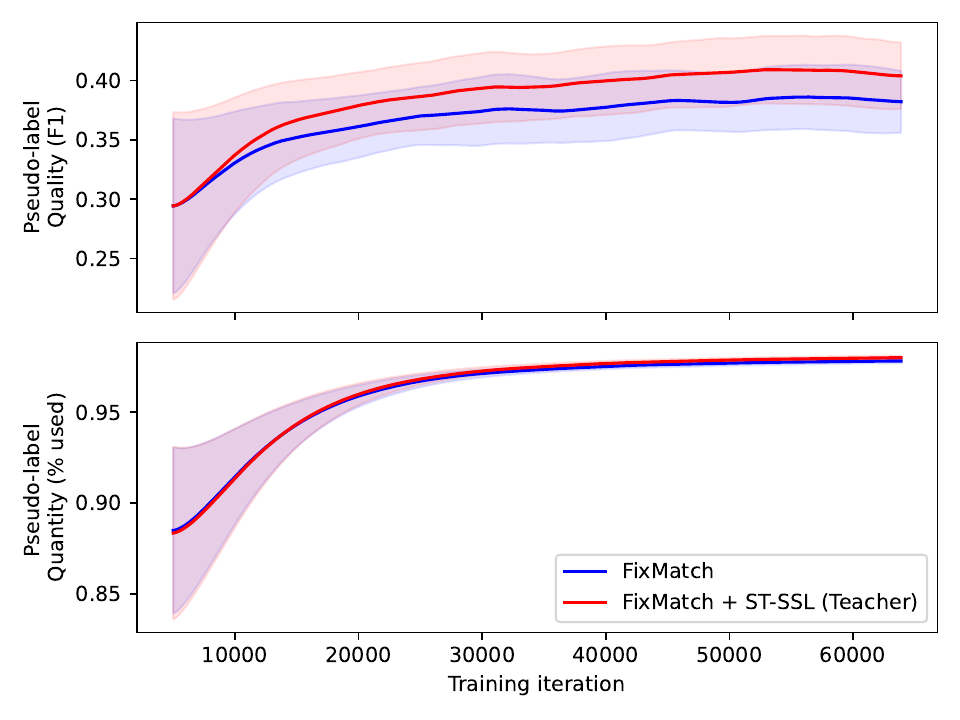}
    \caption{\textbf{Comparing pseudo-label quality (top) and quantity (bottom) of FixMatch and FixMatch + ST-SSL}. The additional metadata in ST-SSL allows generate pseudo-labels of higher quality while the quantity, i.e., the fraction of samples passing FixMatch's confidence threshold, is almost identical. \emph{Curves smoothed for visualization. Best viewed digitally.}}
    \label{fig:pseudo}
    \vspace{5mm}
\end{figure}

    
\begin{table}[t]
    \centering
    \caption{\textbf{Results (accuracy) on EuroSAT.} With few labels, FixMatch + ST-SSL clearly outperforms the baselines, while we see diminishing value of the metadata and ST-SSL for larger numbers of labels.
    }
    \begin{tabular}{ccccc}
    \toprule
    \multirow{2}{*}{\textbf{Method}} & \multicolumn{4}{c}{\textbf{Number of labels}} \\ 
     & \textbf{10} & \textbf{20} & \textbf{40} & \textbf{80} \\
    \midrule
    Supervised only     & 57.56 & 71.70 & 85.13 & 88.50 \\
    FixMatch            & 64.19 & 89.13 & \textbf{94.20} & 96.39 \\
    FixMatch + ST-SSL   & \textbf{77.85} & \textbf{90.56} & 93.96 & \textbf{96.52} \\
    \bottomrule
    \end{tabular}
    \label{tab:eurosat}
    \vspace{0mm}
\end{table}
A central aspect of this work is the assumption that it is not advisable to use metadata inputs at test time as it would have a detrimental effect on the generalization of the model. To investigate this, we analyze different scenarios in Table~\ref{tab:ood}. For this experiment, we train FixMatch + ST-SSL on BigEarthNet with 1\% of training labels and evaluate the performances of the student (not relying on metadata inputs) and the teacher (relying on metadata inputs).

The first row (a) of the table is our final student model as presented in Table~\ref{tab:sota}. The second model (b) corresponds to the teacher model, which relies on the metadata as input at test time. With this model, we observe a slight tendency to overfit on the training samples as the test mAP is 1.19\% lower than the student's. In contrast, the quality of predictions, i.e., pseudo-labels, on the training set is consistently higher than for the non-spatiotemporal counterpart, while the pseudo-label quantity is virtually identical (see Figure~\ref{fig:pseudo}). 

In experiments (c) and (d), we simulated test samples where only the test locations (c) or only the test image acquisition times (d) are outside the training distribution. 
To this end, we select five locations and acquisition times outside the training distribution and replace the true test locations or times with these fixed values (see supplementary material for details).
We report the mean and standard deviation of mAP for the five selected values. When the test locations are outside the training support (c), we observe a huge performance drop of about 15\% for the teacher model. For the acquisition times (d), we observe a significant drop of about 2\%. 
However, the drop is substantially smaller as we encode the acquisition times using the relative day of the year, leaving little space for far out-of-distribution values.
Note that the proposed model (a) is completely unaffected by such distribution shifts in the metadata.

In a practical scenario, however, we may also observe a change in the label and image distribution when the spatiotemporal context changes drastically. 
To investigate this, we create a geographical data split on BigEarthNet, using all samples from Portugal and Ireland only for testing the student (e) and teacher (f). 
In line with our previous results, the student is more robust to the distribution shift than the teacher model. Nonetheless, comparing the order of magnitude of the mAP values for (e,f) with (a-d) indicates that the distributional shifts in the labels and the images have a bigger impact than the distributional shifts in the metadata alone. We leave a deeper investigation to future work as our main concern is SSL and not domain adaption or transfer learning.

Altogether, the takeaway of this set of experiments is that models relying on metadata suffer from inferior generalization to unseen spatiotemporal settings. Therefore, it is desirable to develop models not relying on spatiotemporal metadata, even though the teacher models using the additional input perform better on the training samples and allow to enhance the student's training and performance.

\subsection{EuroSAT}
\label{sec:eurosat}
We also evaluate our approach on EuroSAT, a highly popular benchmark dataset. The results in Table~\ref{tab:eurosat} demonstrate that ST-SSL is able to substantially improve the performance in settings with few labeled samples, e.g., by about 13\% accuracy for ten labels. For 40 and 80 labels, the performances of FixMatch and FixMatch + ST-SSL become similar. We reason that this is because the overall performance of the model is already at a very high level in these settings. Therefore, the additional metadata does not add much value to the model, which is able to classify the vast majority of images correctly anyway.

\subsection{Qualitative Results}
In this section, we qualitatively analyze the effects of the metadata when used as an additional input in our spatiotemporal SSL framework. First, in Figure~\ref{fig:qual}, we illustrate how the metadata improve the predictions on BigEarthNet using the class "sea \& ocean" as an example.
The ST-SSL teacher (Subfigure~(b)) is able to learn the spatial occurrence of the class, which is geographically contiguous in the real world. In concrete terms, this means that the teacher model learns the concept of maritime and continental regions and, thus, mostly avoids false positive predictions in the latter. The learned knowledge is successfully transferred to the student model (Subfigure~(c)), which also mostly avoids this type of error. In contrast, the plain FixMatch model (Subfigure~(a)) produces many false positives in mainland regions, e.g., Eastern Finland. 

To shed more light on this, we conduct an additional experiment where we examine the predicted confidences when various metadata instances are paired with a monochrome, gray image. This allows us to assess the spatiotemporal prior learned by the teacher model. 
In Figure~\ref{fig:all_meta} (a), we can observe high confidences for the class "sea \& ocean" in coastal and maritime regions and low confidences in inland regions on the training dataset (solid regions). On the other hand, the prior is clearly useless on out-of-distribution locations. 
Similarly, we visualize the predicted confidences for "sea \& ocean" depending on the time of image acquisition. We observe a considerable sampling bias toward certain times, which the teacher model can learn. Even though the temporal distribution has no real-world semantic meaning in this case (in contrast to the geospatial distribution of the class), the teacher can exploit it to provide strong pseudo-labels on the training data. The student model benefits from the pseudo-labels, but it cannot take up such a bias as it does not access the metadata.

\section{Conclusion}
In this work, we propose a new SSL framework called \emph{Spatiotemporal SSL}. In this framework, a teacher leverages spatiotemporal metadata to generate high-quality pseudo-labels for a student not receiving the additional input. 
That way, the student can generalize to unseen spatiotemporal contexts while still benefiting from the spatiotemporal information during training. Moreover, we propose a method for joint visual and spatiotemporal modeling and introduce a novel distillation mechanism to enhance the knowledge transfer between teacher and student.
We combine Spatiotemporal SSL with several state-of-the-art SSL algorithms and observe consistent and substantial improvements. 

\paragraph{Limitations}
Even though we have seen consistent improvements with Spatiotemporal SSL, there is a dependency on the information contained in the metadata itself. For instance, in a scenario where the distributions of labels and images are independent of the spatiotemporal context, we cannot expect performance gains with our approach as the metadata then does not convey any useful information for the model.

\paragraph{Generalizations and Future Work}
The proposed paradigm of exploiting additional low-cost data from a second modality to improve SSL is not only applicable to the classification of spatiotemporal images. In fact, it may be beneficial in any other situation where acquiring additional and informative features is substantially easier than annotating numerous samples. For example, within the remote sensing domain, imagery of higher spatial, temporal, or spectral resolution may be collected for a fixed area without large efforts, whereas the high-resolution data may not be available or practicable for use at inference time. Furthermore, the transfer of our approach to other tasks such as object detection and segmentation, or even other domains is conceivable, opening up many directions for future research.

\newpage

\end{document}